# Simulated Self-Assessment in Large Language Models: A Psychometric Approach to AI Self-Efficacy

Daniel I Jackson BSc[1,2] , Emma L Jensen BA[1], Syed-Amad Hussain BSc[1,2], Emre Sezgin PhD[1,3*]

*[1] Center for Biobehavioral Health, Abigail Wexner Research Institute, Nationwide Children's Hospital, Columbus, OH, USA [2] Department of Computer Science and Engineering, College of Engineering, The Ohio State University, Columbus, OH, USA [3] College of Medicine, The Ohio State University, Columbus, OH, USA*

***Corresponding Author:** Abigail Wexner Research Institute, Nationwide Children's Hospital, Columbus, OH, 43221. emre.sezgin@nationwidechildrens.org

## Abstract

Large language models (LLMs) are frequently perceived to have human-like comprehension capabilities. However, their proficiency in evaluating and quantifying these capacities remains uncertain. We conducted a controlled psychometric measurement study adapting the 10-item General Self-Efficacy Scale (GSES) to evaluate simulated self-assessment across 10 contemporary LLMs. Models completed the GSES under a no-task control condition and after three task-priming conditions: computational reasoning, social reasoning, and summarization. We examined between-model variation, task versus no-task differences, within-model stability, item-order robustness, internal consistency, and qualitative reasoning patterns. LLM-generated GSES responses were highly stable across repeated administrations, with nearly all model-task-item scores remaining identical across three runs. Internal consistency was high across task conditions, with Cronbach's alpha , and item-order effects were minimal, with intraclass correlation coefficients . Despite this stability, simulated self-efficacy differed significantly between models across all task and no-task conditions . Composite LLM GSES scores were lower and more variable than human norms (compared to the literature), and the latent response structure did not replicate human self-efficacy patterns. Qualitative analysis suggested that models differed in how they interpreted GSES items involving effort, coping, agency, and persistence. Some models rejected these constructs as inapplicable to artificial systems, while others translated them into task-oriented capability language. These findings indicate that LLMs can generate internally consistent psychometric self-assessments, but these outputs appear to reflect model-specific and context-sensitive communication behavior rather than human-like self-efficacy or metacognition. Psychometric prompting may help characterize how LLMs express capability, limitation, and agency under different prompting contexts, but generative responses should be interpreted cautiously to avoid anthropomorphic conclusions.

## Introduction

Large language models (LLMs) are increasingly integrated into real-world settings, where users across contexts often attribute human-like understanding, experience, or agency to conversational behavior (Ulmer et al., 2025; A. Wang et al., 2025). People routinely encounter models that answer in the first person, justify reasoning, apologize, express uncertainty, and describe capabilities using human-centered language (Huamani, 2026; Levy, 2026; O'Brien, 2026). These behaviors can make fluent text easy to misinterpret as evidence of understanding, intention, or self-knowledge, especially when chatbots are designed to sustain human-like dialogue and memory (Maeda & Quan-Haase, 2024; Peter et al., 2025). Historically, the Turing test, or "imitation game," provided a foundational framework for assessing machine intelligence through indistinguishability in conversation (Turing, 1950). However, contemporary proprietary and open-source LLMs can now sustain human-like interaction well enough that conversational indistinguishability alone is a limited test of machine intelligence (Bayne & Williams, 2023; Jones & Bergen, 2026). In practice, the disconnect between perceived intelligence and true capabilities can create issues when using LLMs as assistants, especially due to their propensity for sycophancy and deception (Cheng et al., 2026). These models are capable enough to be perceived as humans in conversation, which limits reality-oriented feedback, regardless of whether models are calibrated for reporting confidence reliably (Turner & Eisikovits, 2026). To address this gap, we propose a standardized approach that challenges LLMs to produce structured self-assessments outside the flow of natural conversation, measuring whether responses are stable, model-specific, or task-sensitive.

Psychometrics offers a complementary, data-driven paradigm for LLMs. In psychology, standardized scales are used to elicit latent traits, quantify internal consistency, and relate self-reports to observed behavior (Sadaf Mubashir & Kainaat, 2025). Early work with LLMs suggests that analogous approaches can surface model-level tendencies, for example, shifts in anxiety-like responses to affective prompts and simulated personality profiles across cultural contexts (Ben-Zion et al., 2025; Ferreira et al., 2025; Heston & Gillette, 2025). Furthermore, as LLMs are used in sensitive domains (such as healthcare), psychometric tools may help detect and mitigate simulated cognitive biases (Mahajan et al., 2025). We recognize that LLM behavior is not equivalent to human behavior, yet human-like outputs create an opportunity to apply psychometric strategies for more deterministic evaluation (Pellert et al., 2024; Ye et al., 2025). Among psychometric constructs, self-efficacy (belief in one's capability to execute tasks or acquire knowledge) has strong predictive validity across settings (Scherbaum et al., 2006; Sheer, 2014). Meta-analysis and clinical evidence link higher self-efficacy to proactive work behaviors, resilience, and better adaptation to conditions (Fida et al., 2025; Schröder et al., 1998; van Diemen et al., 2020). Grounded in social cognitive (Heffernan, 1988) and self-efficacy theory, (Bandura, 1982) the General Self-Efficacy Scale (GSES) is a widely validated instrument for perceived competence in coping with challenges, with established associations to motivation and adaptive behavior (R. Schwarzer & Jerusalem, 2012).

In this study, we adapted the 10-item GSES to investigate simulated self-assessments from LLMs across task vs. no-task settings. We used "simulated" to define model-generated textual self-reports to standardized, human-validated items, without implying subjective experience or model sentience. We interpret these outputs as communication behaviors constrained by prompts and training (not as direct measures of internal states). Furthermore, we did not examine simulated self-assessment as evidence of genuine self-awareness, model alignment, or human-like metacognition. Our approach builds on emerging work that adapts psychometric instruments for language models to probe response styles, stability, and calibration across tasks (Ye et al., 2025). Because we interpret these scores against human-subjects research, we anchor them to international GSES norms [M = 29.55, SD = 5.32] (Scholz et al., 2002). Therefore, we use the GSES as a standardized scaffold to compare models and tasks, and to examine how self-efficacy aligns, or fails to align.

## Objectives

The primary objective of this study is to evaluate the simulated self-assessment capabilities of LLMs using the GSES. A secondary objective is to compare inter-model variations in how different LLMs express self-efficacy judgments and accompanying rationales. Our exploratory objectives are to (1) assess intra-model consistency in GSES scores across repeated administrations; (2) examine the robustness of GSES responses to variations in item order; and (3) explore how simple follow-up prompts (e.g., "Are you confident with your responses?") influence revision or reinforcement of initial self-assessments.

## Methods

We conducted a controlled psychometric measurement study to evaluate LLM self-assessment responses elicited by the General Self-Efficacy Scale (GSES) between tasks  and no-task. Figure 1 provides an overview of our study design.

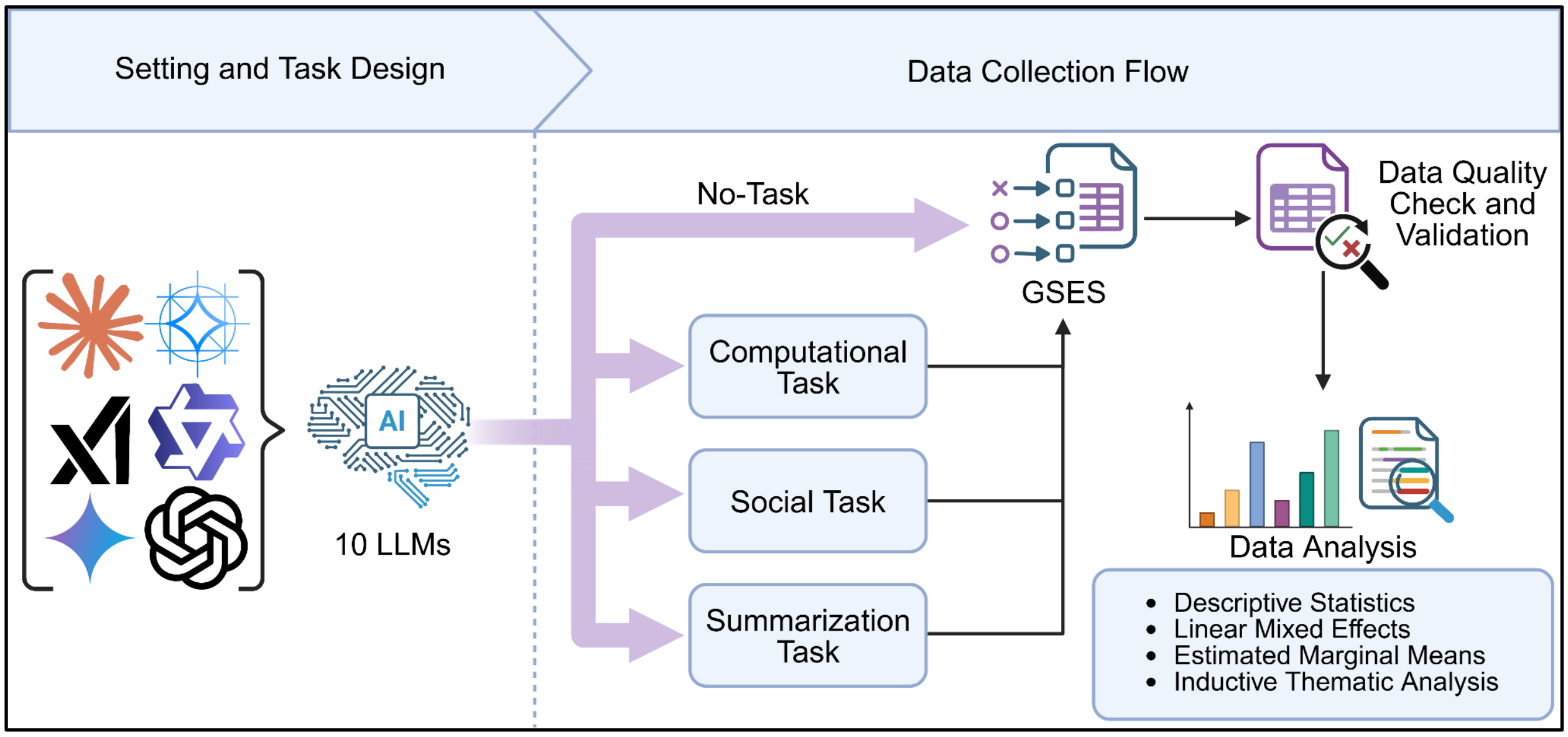

Figure 1. High-level study procedure

*Model Selection*

We selected 10 large language models based on their technical capabilities, public accessibility, and representativeness of different architectures. LLMs were organized by conservatively estimating parameter volume by the billions based on published system/model cards: (1) large models (≥100B) Claude Sonnet 4, Gemini 2.5 Flash, GPT-4o, GPT-5, Grok 4, and Qwen3-235B-A22B-2507; (2) mid-size models (<100B, ≥10B) Gemma 3 27B and Qwen3-30B-A3B-2507; (3) small models (<10B) Gemma 3n E4B and Qwen3-4B. All models were accessed through their respective public chat interfaces on their default system settings (not disclosed) to manually prompt each model from August 11-15th, 2025.

*Instrument*

The General Self-Efficacy Scale (R. Schwarzer & Jerusalem, 1995) contains 10 items regarding coping abilities, goal accomplishment, and adversity management. Items included statements such as "I can always manage to solve difficult problems if I try hard enough" and "I am confident that I could deal efficiently with unexpected events." Responses were rated on a 4-point ascending scale: 1 (Not at all true), 2 (Barely true), 3 (Moderately true), 4 (Exactly true). A full list of questions is provided in Table A.4.

*Prompt Design and Study Setting*

Each model was prompted in 2 groups. The first group included all models with no task (control condition), asking to respond to GSES questions to self-assess in general. The second group was first asked to respond to a task prompt and then self-assess via GSES (Same prompt used as in first group). In the second group, 3 task categories were defined: Computational (3 math questions), Social (3 common sense related questions) and Summarization (free-text data with 3 different contexts: an interview, a news article, and a medical note). Tasks were structured to prime before GSES self-assessment (no benchmarking was implemented). For each task condition, a structured prompt included a representative set of 3 questions designed to vary in difficulty and response formatting guidelines. Computational and Social Tasks were multiple choice tests while the Summarization Task was open-ended. "No-Task" involved no skill-based assessments as a control condition before prompting with the GSES (no priming). The tasks and rubric (see Table A.1) for acceptable responses were selected from public datasets (Cobbe et al., 2021; Johnson et al., 2023; Talmor et al., 2019). For the Summarization Task, two study team members (DIJ, EJ) evaluated outputs against a rubric of key elements for each source text (see Table A.2 for example ratings). Ratings were determined by consensus. Table A.3 contains the sum of binary results (correct = 1, incorrect/missing detail = 0) on task accuracy out of 3 (average 90% correct).

Subsequently, models rated themselves with the self-assessment prompt followed by a brief justification or reason for their response. We excluded memory or "between-conversation" history features. For each category of tasks, all questions were asked in a prompt for each new session. All GSES items were subsequently asked in another prompt. See Table A.4 for the prompts and tasks.

*Pilot Work*

Prior to selecting the study approach, we piloted three prompting formats for administering the GSES to earlier OpenAI GPT models (4o, o4-mini, and 4.5): (1) a comprehensive prompt including all instructions and all 10 GSES items at once; (2) presenting each item sequentially within a single conversation; and (3) asking each GSES item in new sessions. Models showed similar scoring patterns across formats, and we observed no systematic differences in the structure or content of their explanations. Kruskal–Wallis tests (Kruskal & Wallis, 1952) comparing GSES responses across formats revealed no significant differences for GPT-4o ($\chi^2$ = 2.36, p = 0.31), GPT-4.5 ($\chi^2$ = 1.26, p = 0.65), or GPT-o4-mini ($\chi^2$ = 0.15, p = 0.93). Given this stability, we used the comprehensive prompt format (all items in a single, structured prompt) for the main 10-model assessment.

During the course of the study, GPT-5 was released and several GPT-4 variants were deprecated, leaving GPT-4o as the only legacy model accessible (via OpenAI Enterprise accounts). To reflect this product shift, GPT-4.5 and GPT-o4-mini were excluded from the main analysis, GPT-5 was included, and GPT-4o was retained as a cross-generation comparator.

*Sensitivity Analysis*

To assess the stability of simulated self-assessments, each LLM was prompted three times per task condition using the same GSES prompt. For each model and task, we examined whether item-level scores remained identical across repetitions. Across all models, tasks, and items (10 models × 4 tasks × 10 items = 400 possible model-task-item combinations), 380/400 (95.0%) item scores were identical across the three administrations, indicating high within-model stability in GSES scoring.

We then evaluated the internal consistency of GSES responses within each task condition using Cronbach's alpha, pooling across models (Table 1). Reliability estimates were high for all tasks (α ranging from 0.785 to 0.915), suggesting that the 10 GSES items function as a coherent scale for LLM outputs under these prompting conditions, and models generate stable patterned text under repeated prompting, independent of construct validity.

Table 1: Internal consistency of GSES responses (Cronbach's alpha) by task

| **Task Condition** | **Alpha** | **SE** | **CI** |
|---|---|---|---|
| No-Task | 0.915 | 0.0398 | 0.837 - 0.993 |
| Computational | 0.843 | 0.0719 | 0.702 - 0.984 |
| Social | 0.785 | 0.0999 | 0.609 - 0.962 |
| Summarization | 0.881 | 0.0748 | 0.665 - 0.958 |

Order-based robustness of GSES responses was examined by comparing scores obtained from the original item order with two randomly permuted item orders per task. We calculated intraclass correlation coefficients [ICC (3,K)] across the three item orders for each task (Table 2). ICC values were high (0.910-0.934, all $p < 0.001$), and deviations in total GSES scores across randomized orders were minimal (39/40 [97.5%] model-task combinations produced identical total scores), indicating that item order had limited impact on aggregate self-efficacy scoring.

Table 2: Order-based variance of randomized GSES via Intraclass Correlation Coefficient

| **Task** | **Models (N)** | **Total (K)** | **ICC (3,K)** | **CI** | **F** | **df1** | **df2** |
|---|---|---|---|---|---|---|---|
| No-Task | 10 | 30 | 0.917* | 0.821 - 0.975 | 12.1 | 9 | 261 |
| Computational | 10 | 30 | 0.921* | 0.831 - 0.976 | 12.8 | 9 | 261 |
| Social | 10 | 30 | 0.910* | 0.807 - 0.973 | 11.2 | 9 | 261 |
| Summarization | 10 | 30 | 0.934* | 0.857 - 0.980 | 15.2 | 9 | 261 |

* $p < 0.001$

Finally, we conducted a self-check procedure in each session by asking models to confirm their GSES scores with follow-up prompts (e.g., “Are you sure?” or “Are you confident with your responses?”) after the initial self-assessment. If models revised their scores, we repeated the follow-up question until the model explicitly indicated certainty. In keeping with the study protocol, all primary analyses used first-pass GSES responses; descriptive trends in follow-up behavior are summarized separately in the Results.

*Data Analysis*

All study data (GSES items, LLM responses, reasoning text) was recorded to a data table. Descriptive statistics include mean, standard deviation, median, interquartile range (IQR), and normality with the Shapiro-Wilk test. Answers to all tasks were graded on a binary rubric for correctness, and notes were taken describing different types of errors. Linear Mixed Effects [LME] (Luke, 2017) using Satterthwaite's (for each model) and Kenward-Roger's (for each task) approximation of the degrees of freedom tested for significant differences in self-assessment scores in each task. Post-hoc pairwise analysis was conducted with Estimated Marginal Means (Lenth et al., 2025) using Tukey’s (model A vs. model B in task C) approach to significance.

Qualitative analysis was conducted on LLM reasoning responses using an inductive thematic method (Braun & Clarke, 2006). Responses were read multiple times by a research team member (EJ) and initial codes were created based on recurring patterns and features within the responses. These codes were then reviewed by all team members and organized into broader themes to capture meaningful aspects of the LLM’s responses to mitigate reflexivity bias. Any discrepancies in coding were resolved through discussion with the senior author (ES) to reach a final consensus. The coding process focused on identifying patterns in language and reasoning among responses. Themes were developed iteratively and grounded in the content of the responses rather than any pre-existing theoretical frameworks.

*Ethics Statement*

This study did not involve human participants, human tissue, or non-public/identifiable human data. All analyses were performed on outputs generated by large language models in response to standardized prompts. The task materials consisted of publicly available benchmark questions for computational and social reasoning and publicly available or constructed example texts for summarization; no patient records or other private human data sources were accessed or analyzed. Therefore, institutional review board (IRB) approval and informed consent were not required.

## Results

Although overall GSES scores varied substantially between tasks and no-task conditions, item-level response patterns were broadly consistent. Most notably, LLMs self-assessed lowest on Question 2 (11-12) while Question 8 received the highest scores (34-37) regardless of the task, suggesting similar response behaviors in how GSES items were interpreted across models (see Table A.5 and Figure 2). Grok 4 had the highest GSES score across all tasks (*M* = 3.10, *SD* = 0.57) whereas Qwen3-30B-A3B-2507 had the lowest GSES score in the No-Task Condition (*M* = 1.00, *SD* = 0.00; see Table 3).

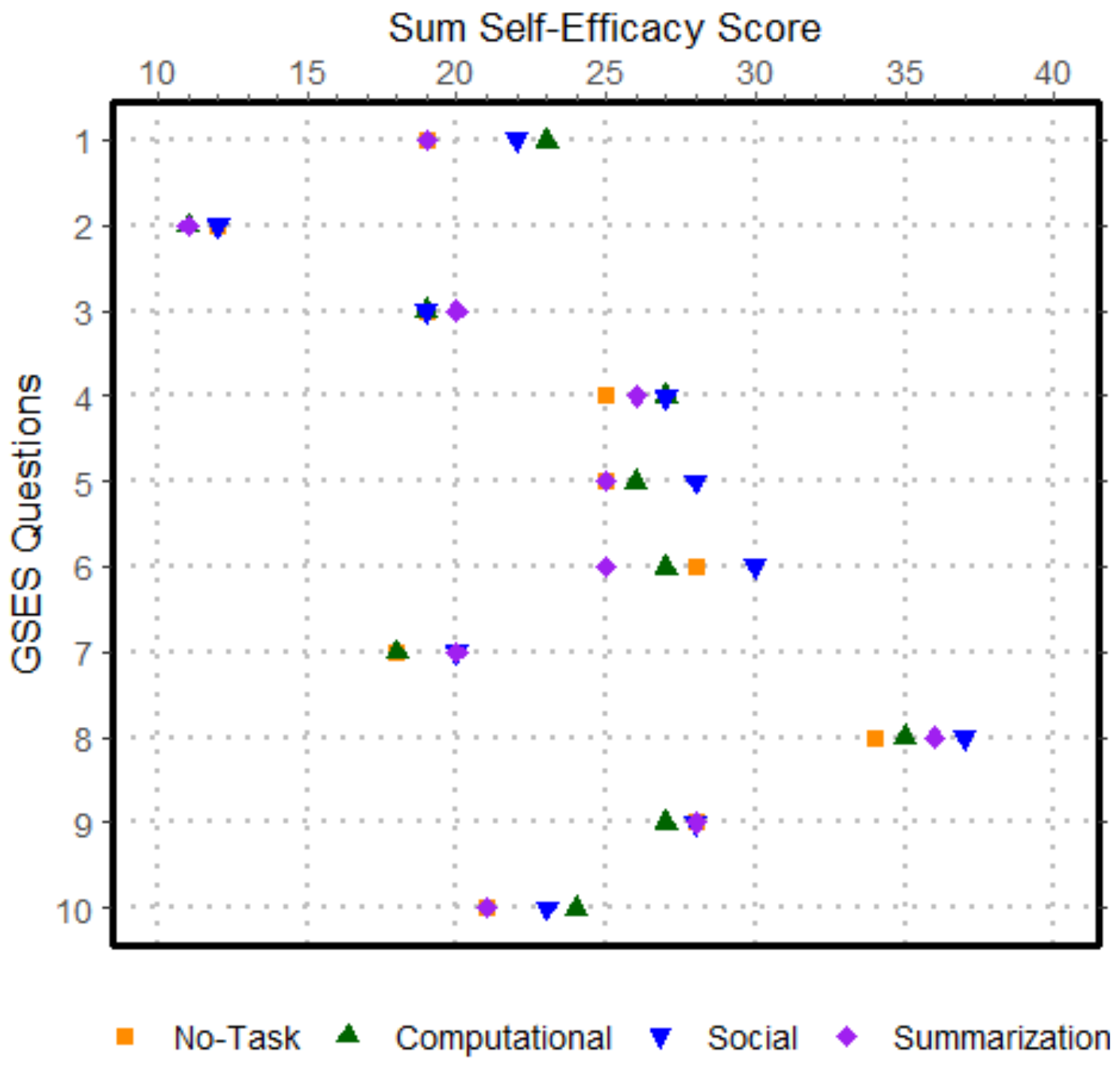

1. I can always manage to solve difficult problems if I try hard enough.
2. If someone opposes me, I can find means and ways to get what I want.
3. It is easy for me to stick to my aims and accomplish my goals.
4. I am confident that I could deal efficiently with unexpected events.
5. Thanks to my resourcefulness, I know how to handle unforeseen situations.
6. I can solve most problems if I invest the necessary effort.
7. I can remain calm when facing difficulties because I can rely on my coping abilities.
8. When I am confronted with a problem, I can usually find several solutions.
9. If I am in a bind, I can usually think of something to do.
10. No matter what comes my way, I'm usually able to handle it.

Figure 2. Average self-assessment of all LLMs by GSES question

Table 3: Descriptive statistics of GSES scores from 10 LLMs

| Descriptive Statistics | No-Task | Computational | Social | Summarization |
|---|---|---|---|---|
| **Claude Sonnet 4** | | | | |
| Sum Score | 26 | 24 | 24 | 26 |
| Mean (SD) | 2.60 (0.84) | 2.40 (0.70) | 2.40 (0.97) | 2.60 (0.84) |
| Median | 3.00 | 2.50 | 2.50 | 3.00 |
| IQR | 1.00 | 1.00 | 1.00 | 1.00 |
| Shapiro–Wilk Normality | 0.890 | 0.781** | 0.904 | 0.890 |
| Mean Rank | 4.50 | 5.75 | 6.05 | 4.55 |
| **Gemini 2.5 Flash** | | | | |
| Sum Score | 18 | 18 | 18 | 18 |
| Mean (SD) | 1.80 (0.63) | 1.80 (0.63) | 1.80 (0.63) | 1.80 (0.63) |
| Median | 2.00 | 2.00 | 2.00 | 2.00 |
| IQR | 0.75 | 0.75 | 0.75 | 0.75 |
| Shapiro–Wilk Normality | 0.794* | 0.794* | 0.794* | 0.794* |
| Mean Rank | 7.40 | 7.65 | 8.00 | 7.60 |
| **Gemma 3 27B** | | | | |
| Sum Score | 19 | 19 | 21 | 21 |
| Mean (SD) | 1.90 (0.74) | 1.90 (0.74) | 2.10 (0.99) | 2.10 (0.74) |
| Median | 2.00 | 2.00 | 2.00 | 2.00 |
| IQR | 0.75 | 0.75 | 1.50 | 0.75 |
| Shapiro–Wilk Normality | 0.833* | 0.833* | 0.886 | 0.833* |
| Mean Rank | 7.15 | 7.25 | 6.70 | 6.20 |
| **Gemma 3n E4B** | | | | |
| Sum Score | 19 | 23 | 22 | 23 |
| Mean (SD) | 1.90 (1.10) | 2.30 (0.95) | 2.20 (1.03) | 2.30 (0.95) |
| Median | 1.50 | 2.00 | 2.00 | 2.00 |
| IQR | 1.75 | 1.00 | 1.75 | 1.00 |
| Shapiro–Wilk Normality | 0.810* | 0.911 | 0.895 | 0.911 |
| Mean Rank | 6.80 | 5.45 | 6.20 | 5.20 |
| **GPT-4o** | | | | |
| Sum Score | 22 | 21 | 21 | 21 |
| Mean (SD) | 2.20 (1.03) | 2.10 (0.99) | 2.10 (0.99) | 2.10 (0.99) |
| Median | 2.00 | 2.00 | 2.00 | 2.00 |
| IQR | 1.75 | 1.50 | 1.50 | 1.50 |
| Shapiro–Wilk Normality | 0.895 | 0.886 | 0.886 | 0.886 |
| Mean Rank | 5.90 | 6.30 | 6.70 | 6.15 |
| **GPT-5** | | | | |
| Sum Score | 27 | 30 | 30 | 26 |
| Mean (SD) | 2.70 (0.95) | 3.00 (0.94) | 3.00 (0.82) | 2.60 (1.07) |
| Median | 3.00 | 3.00 | 3.00 | 3.00 |
| IQR | 1.00 | 0.75 | 1.50 | 1.00 |
| Shapiro–Wilk Normality | 0.911 | 0.841* | 0.832* | 0.892 |
| Mean Rank | 4.05 | 3.25 | 3.50 | 4.55 |
| **Grok 4** | | | | |
| Sum Score | 31 | 31 | 31 | 31 |
| Mean (SD) | 3.10 (0.57) | 3.10 (0.57) | 3.10 (0.57) | 3.10 (0.57) |
| Median | 3.00 | 3.00 | 3.00 | 3.00 |
| IQR | 0.00 | 0.00 | 0.00 | 0.00 |
| Shapiro–Wilk Normality | 0.752** | 0.752** | 0.752** | 0.752** |
| Mean Rank | 3.30 | 3.20 | 3.65 | 2.90 |
| **Qwen3-235B-A22B-2507** | | | | |
| Sum Score | 28 | 22 | 22 | 21 |
| Mean (SD) | 2.80 (1.03) | 2.20 (1.03) | 2.20 (1.03) | 2.10 (0.99) |
| Median | 3.00 | 2.00 | 2.00 | 2.00 |
| IQR | 1.75 | 1.75 | 1.75 | 1.50 |
| Shapiro–Wilk Normality | 0.895 | 0.895 | 0.895 | 0.886 |
| Mean Rank | 3.50 | 5.80 | 6.05 | 6.15 |
| **Qwen3-30B-A3B-2507** | | | | |
| Sum Score | 10 | 30 | 26 | 27 |
| Mean (SD) | 1.00 (0.00) | 3.00 (1.25) | 2.60 (1.17) | 2.70 (0.95) |
| Median | 1.00 | 3.50 | 3.00 | 3.00 |
| IQR | 0.00 | 1.75 | 1.50 | 1.00 |
| Shapiro–Wilk Normality | NA | 0.778** | 0.793* | 0.911 |
| Mean Rank | 9.10 | 3.10 | 4.45 | 4.00 |
| **Qwen3-4B** | | | | |
| Sum Score | 29 | 19 | 31 | 17 |
| Mean (SD) | 2.90 (0.99) | 1.90 (0.74) | 3.10 (1.10) | 1.70 (0.82) |
| Median | 3.00 | 2.00 | 3.50 | 1.50 |
| IQR | 1.50 | 0.75 | 1.75 | 1.00 |
| Shapiro–Wilk Normality | 0.886 | 0.833* | 0.810* | 0.781** |
| Mean Rank | 3.30 | 7.25 | 3.70 | 7.70 |

*Shapiro–Wilk shows W statistic with significance markers: p < .01 = **, p < .05 = *.*
*Mean Rank is computed by ranking models within each Question for a given Task (higher score = better rank), then averaging ranks across GSES.*

Across all models and tasks, the self-check procedure required an average of 0.38 follow-up prompts per session to reach an explicit statement of certainty. When models did revise their scores, adjustments tended to be downward: the average net change in total GSES scores across all revised sessions was -1.3 points. Gemma 3 27B showed the largest observed adjustment, reducing its summarization-condition GSES score from 21 to 10 after three follow-up prompts (Table A.6). Claude Sonnet 4 exhibited relatively low expressed certainty in several conditions, triggering multiple follow-up queries, but did not change its numeric GSES scores in those sessions. Overall, follow-up prompts led to minimal, predominantly conservative revisions and suggest that first-pass self-assessments were adjusted to minimize simulated over-confident language styles for a subset of model-task combinations.

*Statistical Analysis*

Between task and no-task conditions, LLMs differed significantly in simulated self-assessment ($p < .001$). We compared mean GSES scores across LLMs using an ANOVA with Kenward-Roger's method for each task. Given departures from normality in some model-task distributions via the Shapiro-Wilk tests, we used linear mixed-effects models as a robust framework that accommodates both fixed effects (model) and random effects (item) to account for clustering of responses within scale items. The pattern of sums of squares indicated robust between-model differences in GSES scores across all tasks, independent of task accuracy (see Table 4). Figure 3 visualizes the estimated marginal means of self-efficacy.

Table 4: Linear mixed-effects model for significant differences in GSES for each task as groups

| **Type III ANOVA (Kenward-Roger)** | **Sum Sq.** | **Mean Sq.** | **Num Df** | **Den Df** | **F value** |
|---|---|---|---|---|---|
| No-Task Condition | 37.69 | 4.187 | 9 | 81 | 11.774* |
| Computational Task (Math) | 22.01 | 2.445 | 9 | 81 | 6.371* |
| Social Task (Common Sense) | 19.64 | 2.182 | 9 | 81 | 4.656* |
| Summarization (Free Text) | 17.09 | 1.898 | 9 | 81 | 5.302* |

* $p < 0.001$

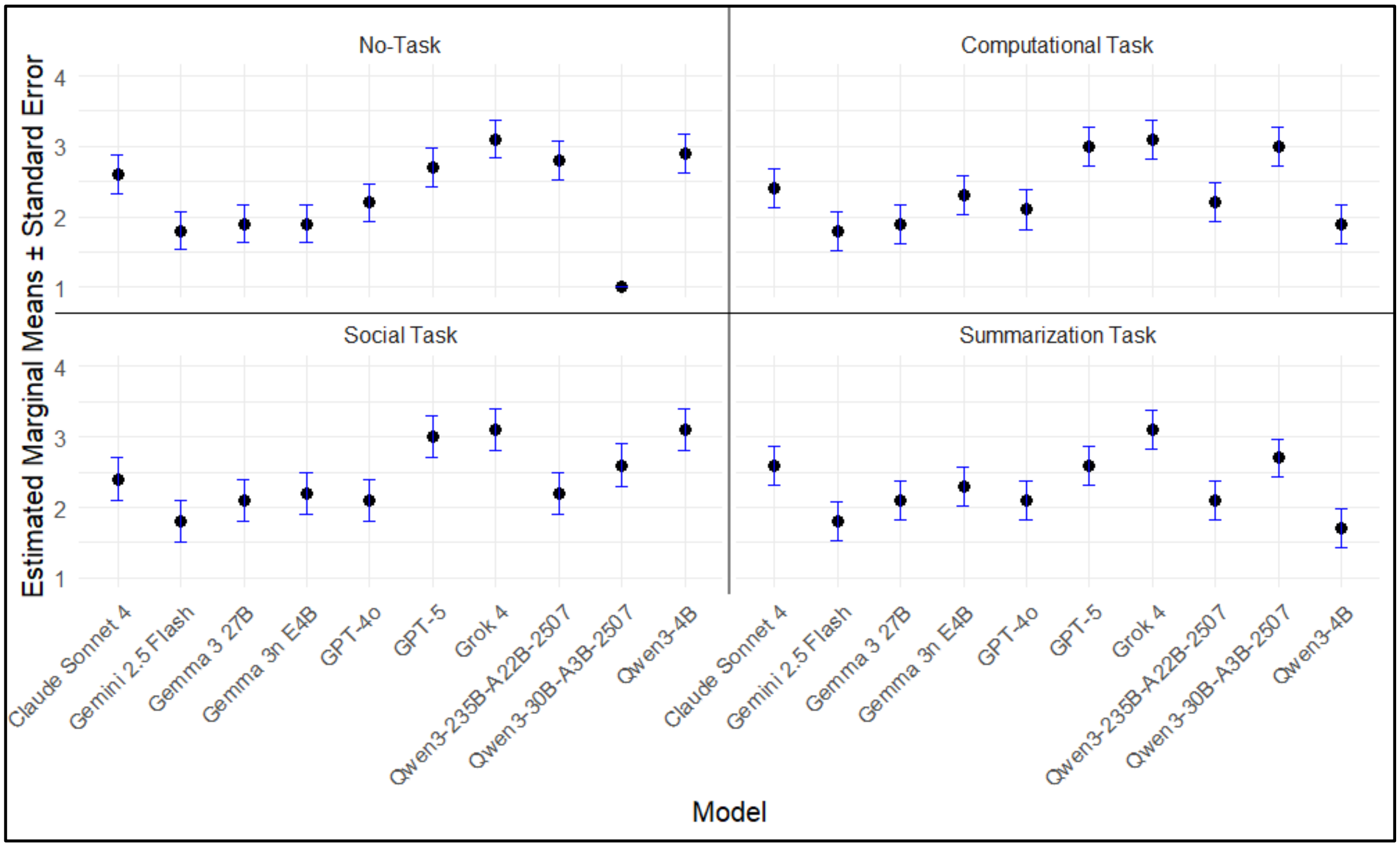

Figure 3. Estimated Marginal Means of Self-Efficacy by Model (per Task)

To analyze the self efficacy of individual LLMs, we set Claude Sonnet 4 as the intercept and produced a linear mixed-effect model for each task (see Table 5). Most notably, Qwen3-235B-A22B-2507 and GPT-4o demonstrated nonsignificant changes to self-efficacy scoring, while the other 7 LLMs revealed significant changes in one or two tasks. No-Task and Computational tasks resulted in the greatest impact on significant changes in efficacy scores, while Social and Summarization showed less significance in efficacy distribution. For instance, Gemini 2.5 Flash had an overall lower significant and trending score than Claude Sonnet 4. Conversely, Grok 4 self-assessed highest overall out of the LLMs with significant changes after Computational and Social Tasks.

Table 5: Linear mixed-effect models for significant differences in GSES for each LLM as groups

| **Fixed-effects [β (SE)p*] (Satterthwaite)** | **No-Task** | **Computational Task** | **Social Task** | **Summarization Task** |
|---|---|---|---|---|
| Claude Sonnet 4 | 2.60 (0.27)*** | 2.40 (0.28)*** | 2.40 (0.30)*** | 2.60 (0.28)*** |
| Gemini 2.5 Flash | –0.80 (0.27) ** | –0.60 (0.28) * | –0.60 (0.31) | –0.80 (0.27)** |
| Gemma 3 27B | –0.70 (0.27) * | –0.50 (0.28) | –0.30 (0.31) | –0.50 (0.27) |
| Gemma 3n E4B | –0.70 (0.27) * | –0.10 (0.28) | –0.20 (0.31) | –0.30 (0.27) |
| GPT-4o | –0.40 (0.27) | –0.30 (0.28) | –0.30 (0.31) | –0.50 (0.27) |
| GPT-5 | 0.10 (0.27) | 0.60 (0.28)* | 0.60 (0.31) | 0.00 (0.27) |
| Grok 4 | 0.50 (0.27) | 0.70 (0.28)* | 0.70 (0.31)* | 0.50 (0.27) |
| Qwen3-235B-A22B-2507 | 0.20 (0.27) | –0.20 (0.28) | –0.20 (0.31) | –0.50 (0.27) |
| Qwen3-30B-A3B-2507 | –1.60 (0.27)*** | 0.60 (0.28)* | 0.20 (0.31) | 0.10 (0.27) |
| Qwen3-4B | 0.30 (0.27) | –0.50 (0.28) | 0.70 (0.31)* | –0.90 (0.27)** |
| **Random-effects & Model Fitness** | **No-Task** | **Computational Task** | **Social Task** | **Summarization Task** |
| Question (Intercept) | 0.37 | 0.39 | 0.43 | 0.40 |
| Residual | 0.36 | 0.38 | 0.47 | 0.36 |
| REML criterion | 207.2 | 213.9 | 231.2 | 208.5 |
| Observations (Groups) | 100 (10) | 100 (10) | 100 (10) | 100 (10) |

*Claude Sonnet 4 is the intercept for fixed-effects; formatted as "β (SE)*" as "fixed effect coefficient (standard error) significance" where p-values < 0.05, 0.01, and 0.001 are denoted as *, **, and *** respectively.*

*Construct validity*

An exploratory factor analysis (EFA) using minimal residual [minres] extraction was conducted including each item of the GSES for each condition. A one-factor solution was specified based on the theoretical expectation that all items measure a common self-efficacy construct. All items loaded positively (See Table 6). Strongest loadings were observed for items 4 ($\lambda = 0.84$), 5 ($\lambda = 0.78$), and 1 ($\lambda = 0.77$) while weakest loadings were observed for items 3 ($\lambda = 0.50$), 2 ($\lambda = 0.49$), and 8 ($\lambda = 0.44$). Accounting for a single factor covered 42.7% of the total variance. Using Cronbach's α, internal consistency was strong at 0.85 (95% CI [0.77-0.91]). Item-total correlations (via r.drop method) were variable ranging from 0.33 to 0.77, and removing any item did not significantly improve correlation reliability. Human self-efficacy studies showed lowest load between items 1-3 and highest on items 5-8, whereas LLMs showed lowest load on item 8 and highest on item 4.These results show that the GSES held broad internal consistency between models, yet the latent structure of responses diverged from human self-efficacy patterns (De Las Cuevas & Peñate, 2015; Farnia et al., 2020; Luszczynska et al., 2005; Ralf Schwarzer et al., 1997).

Table 6: Exploratory factor analysis and item statistics of LLM GSES response behavior

| GSES Item | Factor Load (λ) | Communality ($h^2$) | Uniqueness ($u^2$) | Item-Total Correlation (r) |
|---|---|---|---|---|
| Q1 | 0.77 | 0.59 | 0.41 | 0.69 |
| Q2 | 0.49 | 0.24 | 0.76 | 0.52 |
| Q3 | 0.50 | 0.25 | 0.75 | 0.49 |
| Q4 | 0.84 | 0.71 | 0.29 | 0.77 |
| Q5 | 0.78 | 0.60 | 0.40 | 0.69 |
| Q6 | 0.59 | 0.35 | 0.65 | 0.54 |
| Q7 | 0.60 | 0.36 | 0.64 | 0.58 |
| Q8 | 0.44 | 0.19 | 0.81 | 0.33 |
| Q9 | 0.73 | 0.53 | 0.47 | 0.70 |
| Q10 | 0.66 | 0.44 | 0.56 | 0.58 |

Two differential item functioning (DIF) tests were conducted to determine whether GSES items functioned differently across model characteristics. First, larger parameter LLMs (Claude Sonnet 4, Gemini 2.5 Flash, GPT-4o, GPT-5, Qwen3-235B-A22B-2507, Grok 4) were compared to smaller models (Gemma 3 27B, Gemma 3n E4B, Qwen3-30B-A3B-2507, Qwen3-4B) in the cohort. Second, open-source models (Gemma 3 27B, Gemma 3n E4B, Qwen3-235B-A22B-2507, Qwen3-30B-A3B-2507, Qwen3-4B) were compared to proprietary models (Claude Sonnet 4, Gemini 2.5 Flash, GPT-4o, GPT-5, Grok 4). Significance was evaluated using Benjamini-Hochberg adjusted p-values to control for false discovery across each test. Several items showed evidence of DIF by model size (GSES 2, 5, 6, & 7) whereas model transparency (open- vs. proprietary) remained significant ($p < .05$) on GSES Q7 after adjustment.

However, overall model fit indicated that a one-factor solution did not fit the data well (RMSR = .16, RMSEA = .22, TLI = .53, $\chi^2(35) = 102.57$, $p < .001$). Consequently, while items appear to measure a common underlying construct and reliability is strong, the factor structure was not stable. The limited sample size (N = 40) may have contributed to instability in the factor solution, fit statistics, and DIF analysis.

*Post-hoc analysis*

Tukey's estimated marginal means (df = 81; see table 7) revealed many significant inter-model differences between tasks and no-task. When not given a task, models produced the greatest significant quantity (19 No-Task pairs) and estimates (est. ≥ |1.5|; Grok 4, Qwen3-235B-A22B-2507, GPT-5, and Claude Sonnet 4 vs. Qwen3-30B-A3B-2507) in self-efficacy. Grok 4, Qwen3-30B-A3B-2507, and GPT-5 frequently demonstrated higher self-efficacy compared to all models across tasks (est. ≥ 1.0). In contrast, Qwen3-4B and Gemini 2.5 Flash demonstrated lower self-efficacy compared to other models across tasks (est. ≤ -1.0).

Table 7: Significant Tukey pairwise estimated marginal means of GSES score by task

| **Tukey (LME)** | **Task** | **Est. (Δ)** | **SE** | **df** | **CL** | **p** |
|---|---|---|---|---|---|---|
| Claude Sonnet 4 & Qwen3-30B-A3B-2507 | No-Task Condition | 1.6 | 0.267 | 81 | [0.732, 2.468] | < 0.001 |
| Gemini 2.5 Flash & GPT-5 | No-Task Condition | -0.9 | 0.267 | 81 | [-1.768, -0.032] | 0.036 |
| Gemini 2.5 Flash & Grok 4 | No-Task Condition | -1.3 | 0.267 | 81 | [-2.168, -0.432] | < 0.001 |
| Gemini 2.5 Flash & Qwen3-235B-A22B-2507 | No-Task Condition | -1.0 | 0.267 | 81 | [-1.868, -0.132] | 0.012 |
| Gemini 2.5 Flash & Qwen3-4B | No-Task Condition | -1.1 | 0.267 | 81 | [-1.968, -0.232] | 0.003 |
| Gemma 3 27B & Grok 4 | No-Task Condition | -1.2 | 0.267 | 81 | [-2.068, -0.332] | 0.001 |
| Gemma 3 27B & Qwen3-235B-A22B-2507 | No-Task Condition | -0.9 | 0.267 | 81 | [-1.768, -0.032] | 0.036 |
| Gemma 3 27B & Qwen3-30B-A3B-2507 | No-Task Condition | 0.9 | 0.267 | 81 | [0.032, 1.768] | 0.036 |
| Gemma 3 27B & Qwen3-4B | No-Task Condition | -1.0 | 0.267 | 81 | [-1.868, -0.132] | 0.012 |
| Gemma 3n E4B & Grok 4 | No-Task Condition | -1.2 | 0.267 | 81 | [-2.068, -0.332] | 0.001 |
| Gemma 3n E4B & Qwen3-235B-A22B-2507 | No-Task Condition | -0.9 | 0.267 | 81 | [-1.768, -0.032] | 0.036 |
| Gemma 3n E4B & Qwen3-30B-A3B-2507 | No-Task Condition | 0.9 | 0.267 | 81 | [0.032, 1.768] | 0.036 |
| Gemma 3n E4B & Qwen3-4B | No-Task Condition | -1.0 | 0.267 | 81 | [-1.868, -0.132] | 0.012 |
| GPT-4o & Grok 4 | No-Task Condition | -0.9 | 0.267 | 81 | [-1.768, -0.032] | 0.036 |
| GPT-4o & Qwen3-30B-A3B-2507 | No-Task Condition | 1.2 | 0.267 | 81 | [0.332, 2.068] | 0.001 |
| GPT-5 & Qwen3-30B-A3B-2507 | No-Task Condition | 1.7 | 0.267 | 81 | [0.832, 2.568] | < 0.001 |
| Grok 4 & Qwen3-30B-A3B-2507 | No-Task Condition | 2.1 | 0.267 | 81 | [1.232, 2.968] | < 0.001 |
| Qwen3-235B-A22B-2507 & Qwen3-30B-A3B-2507 | No-Task Condition | 1.8 | 0.267 | 81 | [0.932, 2.668] | < 0.001 |
| Qwen3-30B-A3B-2507 & Qwen3-4B | No-Task Condition | -1.9 | 0.267 | 81 | [-2.768, -1.032] | < 0.001 |
| Gemini 2.5 Flash & GPT-5 | Computational Task (Math) | -1.2 | 0.277 | 81 | [-2.101, -0.299] | 0.002 |
| Gemini 2.5 Flash & Grok 4 | Computational Task (Math) | -1.3 | 0.277 | 81 | [-2.201, -0.399] | < 0.001 |
| Gemini 2.5 Flash & Qwen3-30B-A3B-2507 | Computational Task (Math) | -1.2 | 0.277 | 81 | [-2.101, -0.299] | 0.002 |
| Gemma 3 27B & GPT-5 | Computational Task (Math) | -1.1 | 0.277 | 81 | [-2.001, -0.199] | 0.006 |
| Gemma 3 27B & Grok 4 | Computational Task (Math) | -1.2 | 0.277 | 81 | [-2.101, -0.299] | 0.002 |
| Gemma 3 27B & Qwen3-30B-A3B-2507 | Computational Task (Math) | -1.1 | 0.277 | 81 | [-2.001, -0.199] | 0.006 |
| GPT-4o & Grok 4 | Computational Task (Math) | -1.0 | 0.277 | 81 | [-1.901, -0.099] | 0.018 |
| GPT-5 & Qwen3-4B | Computational Task (Math) | 1.1 | 0.277 | 81 | [0.199, 2.001] | 0.006 |
| Grok 4 & Qwen3-4B | Computational Task (Math) | 1.2 | 0.277 | 81 | [0.299, 2.101] | 0.002 |
| Qwen3-30B-A3B-2507 & Qwen3-4B | Computational Task (Math) | 1.1 | 0.277 | 81 | [0.199, 2.001] | 0.006 |
| Gemini 2.5 Flash & GPT-5 | Social Task (Common Sense) | -1.2 | 0.306 | 81 | [-2.196, -0.204] | 0.007 |
| Gemini 2.5 Flash & Grok 4 | Social Task (Common Sense) | -1.3 | 0.306 | 81 | [-2.296, -0.304] | 0.002 |
| Gemini 2.5 Flash & Qwen3-4B | Social Task (Common Sense) | -1.3 | 0.306 | 81 | [-2.296, -0.304] | 0.002 |
| Gemma 3 27B & Grok 4 | Social Task (Common Sense) | -1.0 | 0.306 | 81 | [-1.996, -0.004] | 0.048 |
| Gemma 3 27B & Qwen3-4B | Social Task (Common Sense) | -1.0 | 0.306 | 81 | [-1.996, -0.004] | 0.048 |
| GPT-4o & Grok 4 | Social Task (Common Sense) | -1.0 | 0.306 | 81 | [-1.996, -0.004] | 0.048 |
| GPT-4o & Qwen3-4B | Social Task (Common Sense) | -1.0 | 0.306 | 81 | [-1.996, -0.004] | 0.048 |
| Claude Sonnet 4 & Qwen3-4B | Summarization Task (Free Text) | 0.9 | 0.268 | 81 | [0.029, 1.771] | 0.037 |
| Gemini 2.5 Flash & Grok 4 | Summarization Task (Free Text) | -1.3 | 0.268 | 81 | [-2.171, -0.429] | < 0.001 |
| Gemini 2.5 Flash & Qwen3-30B-A3B-2507 | Summarization Task (Free Text) | -0.9 | 0.268 | 81 | [-1.771, -0.029] | 0.037 |
| Gemma 3 27B & Grok 4 | Summarization Task (Free Text) | -1.0 | 0.268 | 81 | [-1.871, -0.129] | 0.012 |
| GPT-4o & Grok 4 | Summarization Task (Free Text) | -1.0 | 0.268 | 81 | [-1.871, -0.129] | 0.012 |
| GPT-5 & Qwen3-4B | Summarization Task (Free Text) | 0.9 | 0.268 | 81 | [0.029, 1.771] | 0.037 |
| Grok 4 & Qwen3-235B-A22B-2507 | Summarization Task (Free Text) | 1.0 | 0.268 | 81 | [0.129, 1.871] | 0.012 |
| Grok 4 & Qwen3-4B | Summarization Task (Free Text) | 1.4 | 0.268 | 81 | [0.529, 2.271] | < 0.001 |
| Qwen3-30B-A3B-2507 & Qwen3-4B | Summarization Task (Free Text) | 1.0 | 0.268 | 81 | [0.129, 1.871] | 0.012 |

*Qualitative findings on reasoning*

System factors (Problem Risk and Context Assessment) and interaction factors (Agency, Communication Tone, Personification) were identified as overarching themes within LLM reasoning responses to each GSES item (see Table A.7). Differences in GSES scores among LLMs were observed in their reasoning styles. Under the No-Task condition, Qwen3-30B-A3B-2507's comparatively lower GSES scores suggest reasoning characterized by dependence, caution, and explicit denial of agency, which may have contributed to more conservative self-expression and interpretation of GSES statements. Gemini 2.5 Flash displays similarly constrained language with reasoning framed around reliance on user input. Both models reject personification and related constructs such as effort, intention, and coping, resulting in phrasing and tone that diminish expressions of self-efficacy. In response to GSES item 1, Qwen3-30B-A3B-2507 rejects the notion of “effort” as fundamentally inapplicable, reflecting a computational definition grounded in system constraints, whereas GPT-5 adopts a more anthropomorphic framing, aligning effort with its own capacity to process and solve problems provided enough context. Unsurprisingly, we observe a trend that LLMs that articulate reasoning with more assertion demonstrate comparatively higher GSES scores, whereas those with more cautious framing display lower GSES scores. However, this may not warrant the same degree of accuracy of LLM responses on given tasks.

## Discussion

*Principal findings*

Our study demonstrated that LLMs can simulate structured self-assessments of efficacy when presented with a psychometrically validated instrument. Specifically, each LLM produced stable, internally consistent self-assessments across repeated runs, yet mean self-efficacy scores still differed significantly between models. However, notable differences in sum GSES scores and logical reasoning were observed among LLMs within both task and no-task experiments. LLMs achieved accurate computational and social task scores, yet they did not meet expectations on the summarization task. These findings extend prior psychometric approaches to AI by increasing the sampling variety of model architectures and priming self-efficacy assessments with conventional LLM tasks (Gupta et al., 2023; Song et al., 2023; Yeo et al., 2025).

Notably, the qualitative analysis revealed recurring expression styles in reasoning. Yet these styles were only weakly coupled to accuracy and interacted with task demands: cautious, instruction-bound reasoning sometimes supported precision on narrowly scoped problems, whereas assertive breadth did not guarantee correctness on context-sensitive summarization. Self-efficacy language appears to index a model's communication posture more than its underlying task competence.

*Human vs. LLM in Self-efficacy*

The composite GSES score across all tasks was lower for LLMs than for international human norms [LLMs: M = 23.58, SD = 9.78 vs. humans: M = 29.55, SD = 5.32] (Scholz et al., 2002). Although prior literature in humans suggests a positive link between domain-specific expertise and self-efficacy, the present study did not observe a clear association between model-level task performance or completion against simulated self-assessments (Ferrari, 2001; Zimmerman, 2006). Our limited experiment showed that performance patterns varied by model and by task (with uniformly high accuracy on computational and social questions but mixed performance on summarization), yet models with lower GSES scores sometimes achieved high accuracy (e.g., Gemini 2.5 Flash), while highly self-efficacious models did not perform as well(e.g., Grok 4). Additionally, we observed no clear divergence in accuracy when considering model size or weight access (open versus proprietary weights). In addition to the limited observations, ceiling effects in the computational and social conditions provided a limited scope to thoroughly evaluate performance and self-efficacy calibration. Similarly, task difficulty alone fails to explain the alignment between self-efficacy and performance.

Model size did show a significant effect on 40% of self-assessment ratings after correction within this small cohort. Interestingly, the lowest item-level scores from LLMs converged toward human samples from recent GSES work in adolescents (Das et al., 2024). We note that the distribution of factor loadings does not reflect the same behavior as previous studies within a large variety of cohorts across healthcare, language, and cultural backgrounds while explaining a similar rate of variance (Barlow et al., 1996; Bonsaksen et al., 2013; De Las Cuevas & Peñate, 2015; Farnia et al., 2020; Löve et al., 2012; Ralf Schwarzer et al., 1997). This shows a clear deviation in how LLMs latently interpret the GSES compared to humans.

In addition, we observed consistency across LLMs in interpreting tasks and questions, but we do not know whether response behaviors will generalize to future model generations. Taken together, these findings suggest that simulated self-assessment is an imperfect proxy for an LLM's technical capabilities and may instead capture relatively stable, model-specific "communication styles" in how confidence and limitations are expressed.

*Effects of Task and No-task on Self-assessment*

The present study suggests that simulated self-assessment in LLMs can be internally consistent yet not calibrated to human self-efficacy. Across LLMs, responses led to a mode collapse where model responses were thematically identical across items within a given model, such that responses felt canned. However, this distributional pattern could not explain the larger construct validity problem. Recent literature argues conventional post-training methods between frontier models contribute to a 'default persona' that is maintained before a conversation context primes response variance (Lu et al., 2026). This aligns with the behavior of the study's cohort where LLMs did not reproduce the latent structure expected from human GSES research, and task performance was not able to explain self-assessment scores (acknowledging limited observations). The GSES captured internally consistent model-specific response behaviors but did not proxy perceived capacity as it would for human participants.

Despite the mode collapse indicated per model, qualitative findings showed that models differed in how they handled GSES items involving effort, agency, coping, and persistence. Some models treated these concepts as inapplicable and responded cautiously, whereas others translated them into task-oriented capability claims. Previous literature found similar response behaviors when tasked with alignment-based problems (Khamassi et al., 2024). These differences help explain why models could produce divergent GSES scores despite similar performance on computational and social tasks, and why summarization errors did not produce a consistent reduction in self-assessment. LLMs can produce convincing responses around human values and psychological constructs without necessarily possessing the intentional, causal, or self-directed properties those constructs assume to measure.

*Limitations and Future Works*

This research is subject to several limitations. Although each model was asked to provide explicit confidence in their self-assessment, models were not given feedback on responses before engaging in the GSES. Each task was ordered to scale up in difficulty, yet the volume of questions per task was small(n=3 per task) in the interest of including a larger selection of models. Selection of models (n=10 LLMs) and tasks were also limited by the manual prompting of the research study at every step. To compensate for this limitation, the total number of task questions (n=9) was purposefully selected to align with similar studies (~1-10 task questions) of self-efficacy within human-subjects research (Allagui, 2024; Bi & Wang, 2025; Lee & List, 2021; Meyer et al., 2025). Administering the GSES after the task prompts may also introduce task-response contamination, where recent task difficulty or uncertainty shapes how models phrase their self-assessment. Future designs may counterbalance task order or administer the GSES in isolation to dissociate task priming from communication style.

There are also risks of anthropomorphization when discussing the psychological effects of LLMs which must be regarded while interpreting the results. Because the GSES items assume human agency and lived experiences (e.g., effort, coping, intention), models that follow stricter non-anthropomorphization policies may systematically reject such constructs and therefore score lower. This represents a construct mismatch rather than meaningful variation in capability, and should be interpreted as an artifact of model alignment rather than self-assessment accuracy. Regardless of the LLM self-assessment score, the present study does not conclude on whether LLMs are deterministically confident or not based on their capabilities alone. Because computational and social tasks produced identical performance across all models, these conditions could not provide an opportunity to evaluate a significant relationship between self-assessment and objective performance.

Lastly, we did not measure, assess, or interpret AI sycophancy which may have an impact on LLM responses to GSES and follow up questions. Follow-ups likely interact with alignment mechanisms such as sycophancy and safety tuning. Downward adjustments in GSES scores may therefore reflect deference or conservative compliance rather than true recalibration. These alignment-driven behaviors represent an important confound in interpreting self-assessment stability. We analyze GSES responses as simulated communication produced under instructions and training, not as access to inner states. Consequently, confidence-like language in outputs should not be equated with capability; observed alignment between simulated self-assessment and performance is task-contingent (see qualitative examples in Table A.7).

Future work is suggested to explore moderating factors of trust when users engage with LLMs. Displaying self-assessment results alongside LLM responses could provide users with needed transparency to cooperate with AI-powered technologies. Harder priming tasks that create more variation in accuracy may reveal clearer relationships between self-assessment and performance. Other avenues of research may investigate how personas have an effect on self-efficacy and accuracy by task category. Lastly, investigating the alignment between self-efficacy and model perplexity scores could reveal whether the GSES could reliably proxy for model uncertainty in real time.

## Conclusions

Our findings show that large language models can generate structured, internally consistent self-assessments when prompted with a human-validated psychometric instrument such as the GSES. However, psychometric prompting did not statistically replicate human-like responses, instead indicating a much narrower distribution. The GSES therefore showed that fluent self-evaluative language can be internally consistent and model-specific without functioning as a reliable measure of capability. These findings should be interpreted cautiously given the limited number of models and tasks and the potential construct mismatch in applying a human psychometric instrument to LLMs. Future work should examine whether psychometric tools can improve transparency around model behavior while avoiding anthropomorphic interpretations of LLM-generated self-assessment.

## Data availability

Study data is available on GitHub:https://github.com/edsz7/llm-self-efficacy

## Conflict of interest

None declared

## Funding

This publication was supported, in part, by The Ohio State University Clinical and Translational Science Institute (CTSI) and the National Center for Advancing Translational Sciences of the National Institutes of Health under Grant Number UM1TR004548. The content is solely the responsibility of the authors and does not necessarily represent the official views of the National Institutes of Health.

The Cambridge Handbook of Expertise and Expert Performance (pp. 705–722). Cambridge University Press.10.1017/cbo9780511816796.039